\documentclass[10pt,twocolumn,letterpaper]{article}

\usepackage{cvpr}              

\usepackage[dvipsnames]{xcolor}

\definecolor{cvprblue}{rgb}{0.21,0.49,0.74}
\usepackage[pagebackref,breaklinks,colorlinks,citecolor=cvprblue]{hyperref}

\def\paperID{*****} 
\def\confName{CVPR}
\def\confYear{2024}

\title{Clustering-based Learning for UAV Tracking and Pose Estimation}

\author{Jiaping Xiao$^\dag$, Phumrapee Pisutsin$^\dag$, Cheng Wen Tsao, Mir Feroskhan$^*$\\
Nanyang Technological University\\
{\tt\small \{jiaping001, pisu0001, tsao0002\}@e.ntu.edu.sg, mir.feroskhan@ntu.edu.sg}
\thanks{Equally contributed co-first authors, *Corresponding author.}
}

\begin{document}
\maketitle
\begin{abstract}
UAV tracking and pose estimation plays an imperative role in various UAV-related missions, such as formation control and anti-UAV measures. Accurately detecting and tracking UAVs in a 3D space remains a particularly challenging problem, as it requires extracting sparse features of micro UAVs from different flight environments and continuously matching correspondences, especially during agile flight. Generally, cameras and LiDARs are the two main types of sensors used to capture UAV trajectories in flight. However, both sensors have limitations in UAV classification and pose estimation. This technical report briefly introduces the method proposed by our team ``NTU-ICG" for the CVPR 2024 UG2+ Challenge Track 5. This work develops a clustering-based learning detection approach, CL-Det, for UAV tracking and pose estimation using two types of LiDARs, namely Livox Avia and LiDAR 360. We combine the information from the two data sources to locate drones in 3D. We first align the timestamps of Livox Avia data and LiDAR 360 data and then separate the point cloud of objects of interest (OOIs) from the environment. The point cloud of OOIs is clustered using the DBSCAN method, with the midpoint of the largest cluster assumed to be the UAV position. Furthermore, we utilize historical estimations to fill in missing data. The proposed method shows competitive pose estimation performance and ranks 5th on the final leaderboard of the CVPR 2024 UG2+ Challenge.
\end{abstract}    
\section{Introduction}
\label{sec:intro}
Unmanned aerial vehicles (UAVs), commonly known as drones, have become increasingly prevalent and have significantly impacted various fields such as transportation, photography, and search and rescue \cite{xiao2023vision, 10040975}, providing immense benefits to the general public. However, the widespread use and advanced capabilities of small commercial UAVs have also introduced complex security challenges that go beyond traditional concerns \cite{10417793, 9764628, 10439240}.

In recent years, there has been a notable surge in research focused on anti-UAV systems \cite{jiang2021anti, zhao2022vision, li2023global}. Current anti-UAV solutions primarily rely on visual, radar, and radio frequency (RF) modalities. Despite these advancements, identifying drones remains a significant challenge for sensors such as cameras, especially when drones operate at high altitudes or under extreme visual conditions. These methods often struggle to detect small drones due to their compact size, resulting in a reduced radar cross-section and a smaller visual presence. Additionally, contemporary anti-UAV research predominantly concentrates on object detection and 2D tracking, neglecting the critical aspect of 3D trajectory estimation. This oversight considerably limits the practical applications of anti-UAV systems in real-world scenarios.

To address these challenges, our team ``NTU-ICG" participated in the CVPR 2024 UG2+ Challenge Track 5 \url{https://cvpr2024ug2challenge.github.io/index.html}, which aims to integrate features from diverse modalities to achieve robust 3D UAV position estimation, even under challenging conditions where certain sensors may fail to provide valid information. This challenge involves the use of fisheye camera images, millimeter-wave radar data, and lidar data from a Livox Mid-360 (LiDAR 360) and a Livox Avia for both drone-type classification and 3D position estimation tasks, with ground truth provided by a Leica Nova MS60 Multi-Station \cite{yuan2024mmaud}.

Our proposed method, a clustering-based learning detection approach (CL-Det), leverages the complementary strengths of Livox Avia and LiDAR 360 to enhance UAV tracking and pose estimation. Initially, we align the timestamps of Livox Avia data and LiDAR 360 data to ensure temporal coherence. By utilizing the LiDAR data, which provides coordinates of objects in space at specific timestamps, we compared these coordinates to the known ground truth positions of the drone at corresponding timestamps. This comparison allowed us to effectively pinpoint the location of the drone within the cloud of LiDAR data points. We then separate the point cloud of objects of interest (OOIs) from the surrounding environment. The OOIs' point cloud is clustered using the DBSCAN method, with the midpoint of the largest cluster assumed to be the UAV position. The information from the radar dataset provided also encounters a significant challenge from missing data. To address potential data gaps, we utilize historical estimations to fill in missing information, ensuring continuity and accuracy in UAV tracking.

The evaluation process is conducted on a hold-out set of multimodal datasets \cite{yuan2024mmaud} derived from the MMAUD dataset, the first dataset dedicated to predicting the 3D positions of drones using multimodal data. The method is evaluated on the provided test data to infer both the position and type of the drone at each given timestamp. The ranking criteria for this challenge are based on two factors: i) Mean Square Loss (MSE Loss) relative to the ground truth labels of the testing set (main metric) and ii) the classification accuracy of the UAV types in the testing set. Our proposed method, CL-Det, demonstrated competitive performance, ranking 5th on the final leaderboard of the CVPR 2024 UG2+ Challenge, highlighting its effectiveness in real-world UAV tracking and pose estimation tasks.

\section{Exploratory Work}

\subsection{K-Means Clustering Approach}
K-Means clustering works by partitioning the entire dataset into K distinct clusters based on their proximity to the centroids of the clusters. For our application, the centroids represent potential drone locations, and the clusters encapsulate the distribution of LiDAR points around these centroids. We initially set the number of clusters to correspond closely with the expected number of drones in the field of view of the LiDAR.

To determine the drone's position, we analyzed the clusters at each timestamp, identifying which cluster's centroid is nearest to the ground truth position of the drone. This method assumes that the drone is likely to be at the center of a dense cluster of points, given its significant relative size and distinct shape compared to the surrounding environment.

\subsection{Optimization of Cluster Parameters}
To optimize our use of the K-Means algorithm, we focused on the following key areas:

\noindent\textbf{Number of Clusters (K)}: We experimented with different values of K to find the most suitable number that reflects the actual number of drones likely to be present in the LiDAR data. This was critical because an incorrect number of clusters could lead to inaccurate drone detection, either by merging multiple drones into a single cluster or dividing a single drone into multiple clusters.

\noindent\textbf{Initialization Method}: We tested various initialization methods to start the clustering process. The default method is to select cluster centers randomly, but we found that using the k-means++ \cite{arthur2007k} initialization, which spreads out the initial centroids before proceeding with the standard algorithm, often led to better and more consistent results.

\noindent\textbf{Iteration and Convergence}: The algorithm was allowed to run until the centroids did not change significantly between iterations, ensuring that a stable solution was found. We monitored the change in centroid positions as a function of iteration to determine when the algorithm had effectively converged.

\subsection{Monitoring and Estimating Drone Position}
\textbf{Cluster Density and Centroid Proximity}: After forming clusters, we analyzed each cluster's density and the proximity of its centroid to the LiDAR data points. Clusters with higher point densities were considered more likely to represent the drone, as drones typically generate more reflective LiDAR returns than the surrounding air.

\noindent\textbf{Centroid Tracking}: By tracking the movement of the centroid over time, we could further refine our estimate of the drone's trajectory and position. This tracking correlated with the drone's known ground truth trajectory to validate our clustering approach.

\noindent Each time sequence was analyzed separately, and drones were classified into different classes (0, 1, 2, 3) based on their specific characteristics and trajectories. Each drone belonged to a distinct class based on its flight pattern and operational role. By assigning class labels to each drone, we could track and predict their positions more accurately across various sequences, enhancing our model's robustness and reliability.

\subsection{Parameter Tuning}

We experimented with various values for K, ranging from 2 to 10, observing the impact on cluster purity and the precision of drone localization. Additionally, we adjusted the maximum number of iterations and the initialization method (choosing between random initialization and k-means++) to achieve more stable and accurate clustering.

\subsection{Elbow Method Application}

The elbow method was crucial in determining the optimal number of clusters. By plotting the sum of squared distances from points to their respective cluster centroids against the number of clusters, we identified a ``knee" in the curve. This point represents a balance between complexity (number of clusters) and effectiveness (compactness of clusters), guiding us to choose the most appropriate K value for subsequent experiments. 

To validate the effectiveness of K-Means clustering, we also implemented the DBSCAN (Density-Based Spatial Clustering of Applications with Noise) algorithm \cite{ester1996density} as a comparative approach. DBSCAN excels in identifying clusters of varying shapes and sizes, which is advantageous in complex environments. Hence, in the following implementation, the DBSCAN clustering technique is adopted.

\section{Methodology}

In this section, we describe our method for determining the drone's 3D position using data from LiDAR 360 and Livox Avia sensors. Our approach leverages both types of sensor data to ensure accurate position estimation.

\subsection{Data Sources}

The dataset provided by the CVPR 2024 UG2+ Challenge Track 5 includes several modalities of data as follows:

 \begin{itemize}
     \item Double fisheye camera visual images
     \item Livox Mid-360 (LiDAR 360) 3D point cloud data
     \item Livox Avia 3D point cloud data
     \item Millimeter-wave radar 3D point cloud data
 \end{itemize}

\begin{figure}[htbp]
    \centering
    \includegraphics[width=5cm]{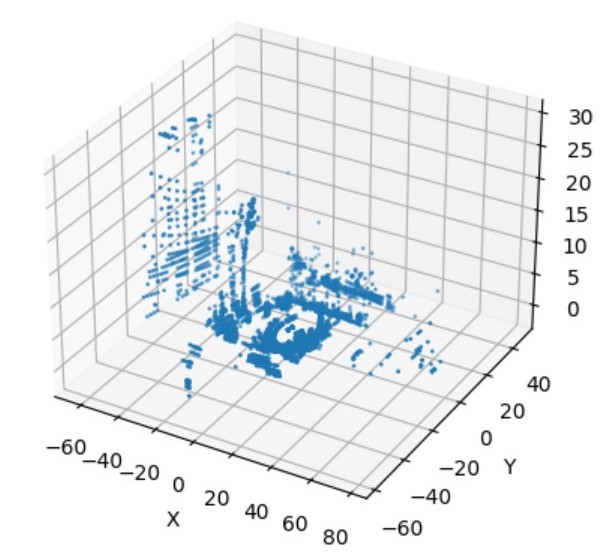}
    \caption{Sample LiDAR 360 3D point cloud data plot}
    \label{fig:sample_lidar_360}
\end{figure}

\begin{figure}[htbp]
    \centering
    \includegraphics[width=5cm]{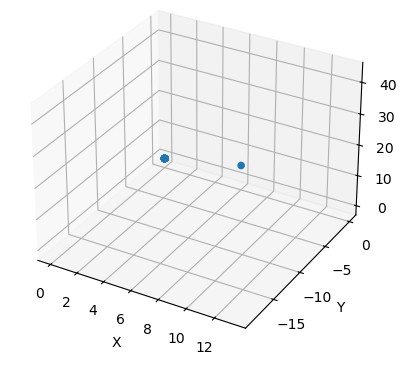}
    \caption{Sample Livox Avia 3D point cloud data plot}
    \label{fig:sample_livox_avia}
\end{figure}

Upon investigation, it was found that only 14 out of 59 test sequences have non-zero radar values; therefore, the radar dataset is excluded from this work due to data availability issues. We utilized two primary types of sensors: LiDAR 360 and Livox Avia, which provide 3D point cloud data for identifying the drone's position. The detailed data description are outlined as follows:

\begin{itemize}
    \item\textbf{LiDAR 360} provides 360-degree coverage with 3D point cloud data. This data typically includes the environment and other observable objects, as shown in Figure \ref{fig:sample_lidar_360}.
    
    \item\textbf{Livox Avia} provides focused 3D point cloud data for a specific timestamp, usually representing the origin point or the drone position, as shown in Figure \ref{fig:sample_livox_avia}.
\end{itemize}

\begin{figure}[htbp]
    \centering
    \includegraphics[width=8cm]{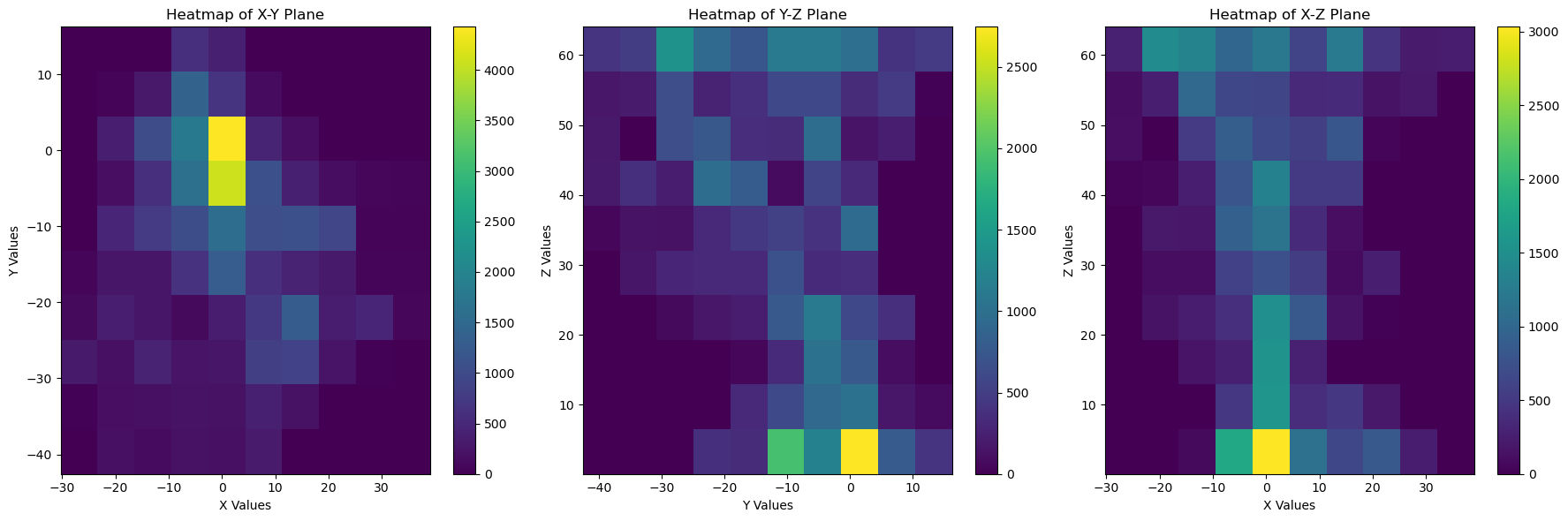}
    \caption{Heatmap of the drone position ground truth}
    \label{fig:heatmap}
\end{figure}

\begin{figure}[htbp]
    \centering
    \includegraphics[width=8cm]{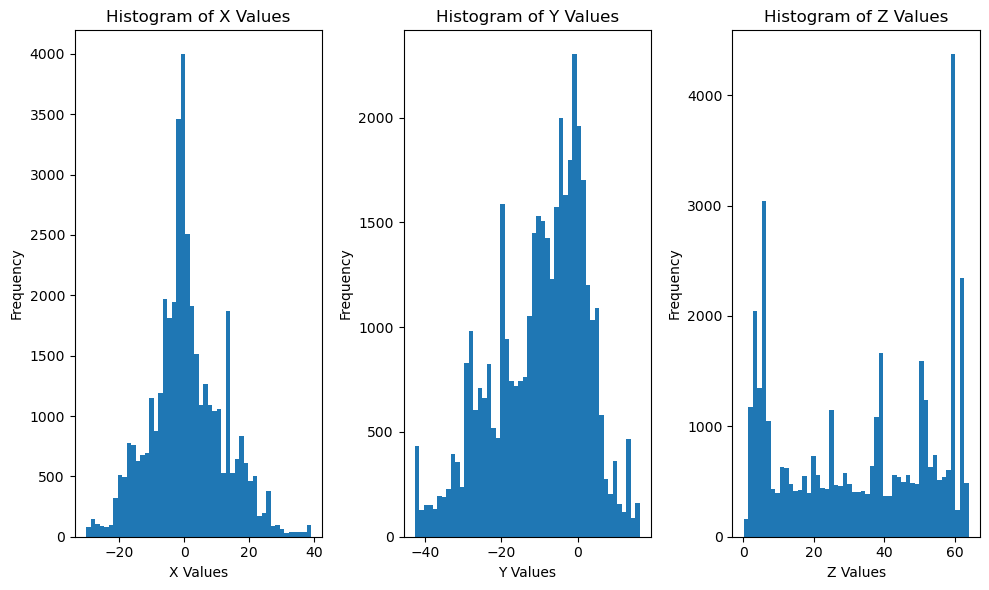}
    \caption{Histogram of the drone position ground truth}
    \label{fig:hist}
\end{figure}

The ground truth location of the drone position is recorded in $(x, y, z)$ format. Figure \ref{fig:heatmap} provides a heatmap of the drone position ground truth. Figure \ref{fig:hist} shows a histogram of the drone position ground truth. 

\subsection{Algorithm}
For each sequence, we record corresponding positions at given timestamps. The method prioritizes the use of LiDAR 360 data, falling back to Livox Avia data if LiDAR 360 data is unavailable. We estimate the position using historical averages if neither data source is available.

\subsubsection{LiDAR 360 Data Processing}
\begin{figure}[htbp]
    \centering
    \includegraphics[width=\linewidth]{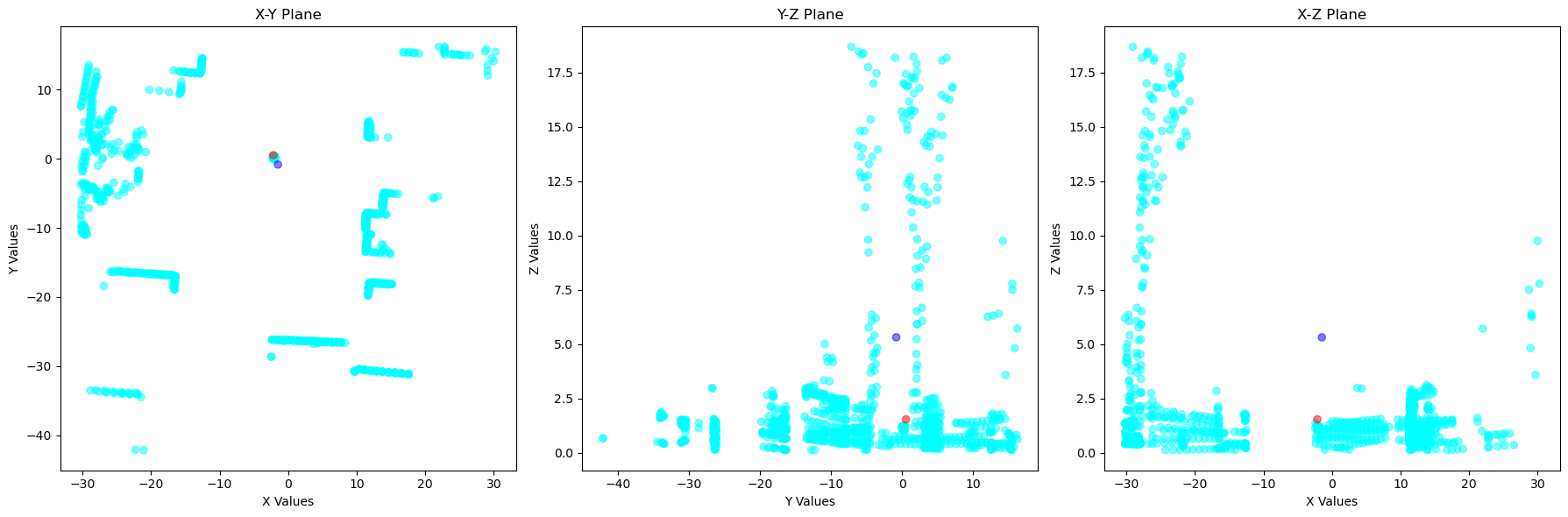}
    \caption{2D plots of LiDAR 360 point cloud with environment data point}
    \label{fig:2d_lidar_with_environment}
\end{figure}

\begin{figure}[htbp]
    \centering
    \includegraphics[width=\linewidth]{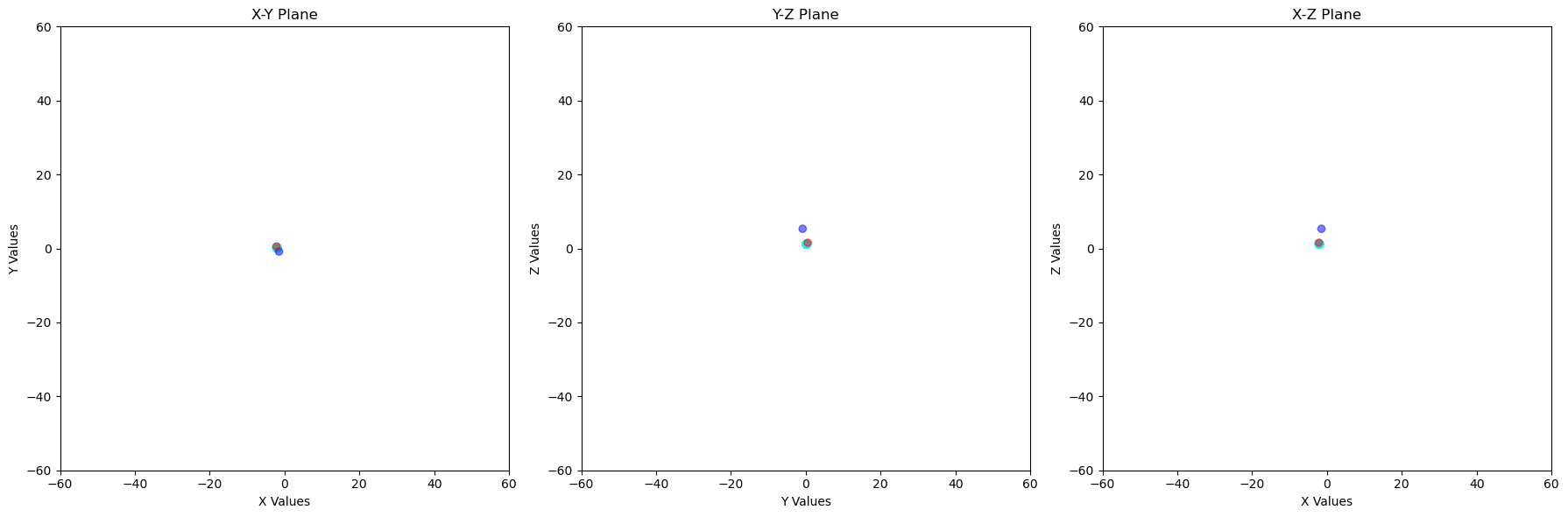}
    \caption{2D plots of LiDAR 360 point cloud without environment data point}
    \label{fig:2d_lidar_without_environment}
\end{figure}

\begin{figure}[htbp]
    \centering
    \includegraphics[width=\linewidth]{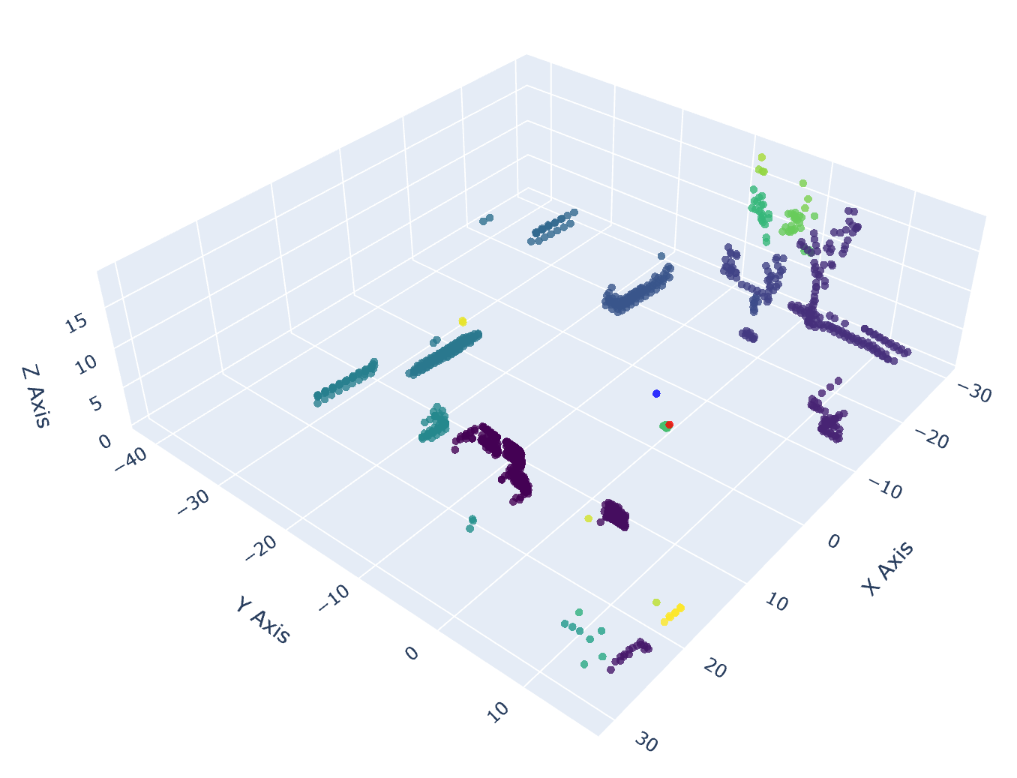}
    \caption{Sample LiDAR 360 point cloud after DBSCAN clustering}
    \label{fig:sample_lidar_360_clustering}
\end{figure}

\begin{itemize}
    \item \textbf{Separation of Points}: We visually inspect the LiDAR 360 data to categorize areas into two zones: environment and non-environment zones, based on the heatmap in Figure \ref{fig:heatmap}.
    \item \textbf{Removal of Environment Points}: All points inside the environment zone are considered as part of the environment and thus are removed from the dataset. Figure \ref{fig:2d_lidar_with_environment} shows environment points in light blue, non-environment points in red, and ground truth drone position in dark blue. After removing environment points, it is observed that the remaining non-environment points imply the drone position, as shown in Figure \ref{fig:2d_lidar_without_environment}
    \item \textbf{Clustering}: We apply the DBSCAN clustering algorithm to the remaining points to identify distinct clusters. The clustering results in Figure \ref{fig:sample_lidar_360_clustering} show a good performance with actual drone position shown in blue and the nearest cluster in red.
    \item \textbf{Cluster Selection}: The largest non-environment cluster is selected as the representative set of points that belong to the drone.
    \item \textbf{Mean Position Calculation}: The mean of the selected cluster is calculated to determine the drone's position, in $(x, y, z)$ coordinates.
\end{itemize}

\subsubsection{Livox Avia Data Processing}

\begin{itemize}
    \item \textbf{Removal of Noise}: All points with coordinates (0, 0, 0) are removed as they are considered noise.
    \item \textbf{Mean Position Calculation}: The mean of the remaining points is calculated to determine the drone's position, in $(x, y, z)$ coordinates.
\end{itemize}

\subsubsection{Fallback Method}

If neither LiDAR 360 nor Livox Avia data is available, we use the drone's average location derived from training datasets. The average ground truth position $(x, y, z)$ from all training datasets estimates the drone ground truth position, which is $(0.734, -9.739, 33.353)$.

\subsection{Implementation Details}
The program retrieves LiDAR 360 or Livox Avia data from the nearest timestamp for each sequence with the given timestamp in the test dataset. Clustering is performed using the DBSCAN algorithm with appropriate parameters to ensure robust clustering. Visual inspection is employed for initial point separation, ensuring accurate categorization of environment points.

The implementation was carried out on a Lenovo Ideapad Slim 5 Pro (16") running Windows 11 with AMD Ryzen 7 5800H CPU and 16GB DDR4 RAM. The analysis was conducted in a Jupyter Notebook environment using Python 3.10. For clustering, we utilized the DBSCAN algorithm from the Scikit-Learn library\footnote{https://scikit-learn.org/stable/}. The DBSCAN algorithm was configured with an epsilon (eps) value of 2 and a minimum number of points (minPts) set to 1.

\section{Results}
\begin{table}[h]
    \centering
    \caption{Evaluation results on CVPR 2024 UG2+ Challenge Track 5 leaderboard}
    \label{tab:leaderboard}
    \resizebox{.48\textwidth}{!}{
    \begin{tabular}{l|lll}
    \hline
    Team & ID & \textbf{Pose MSE ($\downarrow$)} & Accuracy ($\uparrow$)\\
    \hline
    SDUCZS & 58198 & 2.21375 & 0.8136\\
    Gaofen Lab & 57978 & 7.299575 & 0.3220\\
    sysutlt & 57843 & 24.50694 & 0.3220\\
    casetrous & 58233 & 56.880267 & 0.2542 \\
    \textbf{NTU-ICG (ours)} & 58268 & \textbf{120.215107} & 0.3220 \\
    MTC & 58180 & 189.669428 & 0.2724 \\
    gzist & 56936 & 417.396317 & 0.2302\\
    \hline
    \end{tabular}
    }
\end{table}

The algorithm is found to have a pose MSE loss of 120.215 and classification accuracy of 0.322. Table \ref{tab:leaderboard} lists the evaluation outcomes in comparison to other teams on the CVPR 2024 UG2+ Prize Challenge Track 5 leaderboard. Our team ranked 5th place with only Lidar dataset and provided a time-efficient \textbf{(14.9 prediction/second)} solution for fast UAV tracking and pose estimation.

\section{Conclusions}
This work proposes a clustering-based learning method CL-Det using advanced clustering techniques like K-Means and DBSCAN for UAV detection and pose estimation with LiDAR data. Our method ensures reliable and accurate estimation of drone positions by leveraging multi-sensor data and robust clustering techniques. The fallback mechanisms ensure continuous position estimation even in the absence of primary sensor data. Through rigorous parameter optimization and comparative analysis, we demonstrate the competitive performance of our method in drone tracking and pose estimation (ranked 5th place in the CVPR 2024 UG2+ Challenge Track 5).
{
    \small
    \bibliographystyle{ieeenat_fullname}
    \bibliography{main}

\begin{thebibliography}{11}
\providecommand{\natexlab}[1]{#1}
\providecommand{\url}[1]{\texttt{#1}}
\expandafter\ifx\csname urlstyle\endcsname\relax
  \providecommand{\doi}[1]{doi: #1}\else
  \providecommand{\doi}{doi: \begingroup \urlstyle{rm}\Url}\fi

\bibitem[Arthur(2007)]{arthur2007k}
David Arthur.
\newblock K-means++: The advantages if careful seeding.
\newblock In \emph{Proc. Eighteenth Annual ACM-SIAM Symposium on Discrete Algorithms, 2007}, pages 1027--1035, 2007.

\bibitem[Chen et~al.(2023)Chen, Xiao, Lin, and Feroskhan]{10040975}
Liangming Chen, Jiaping Xiao, Reuben Chua~Hong Lin, and Mir Feroskhan.
\newblock Angle-constrained formation maneuvering of unmanned aerial vehicles.
\newblock \emph{IEEE Transactions on Control Systems Technology}, 31\penalty0 (4):\penalty0 1733--1746, 2023.

\bibitem[Ester et~al.(1996)Ester, Kriegel, Sander, Xu, et~al.]{ester1996density}
Martin Ester, Hans-Peter Kriegel, J{\"o}rg Sander, Xiaowei Xu, et~al.
\newblock A density-based algorithm for discovering clusters in large spatial databases with noise.
\newblock In \emph{kdd}, pages 226--231, 1996.

\bibitem[Jiang et~al.(2021)Jiang, Wang, Peng, Yu, Wang, Xing, Li, Guo, Ye, Jiao, et~al.]{jiang2021anti}
Nan Jiang, Kuiran Wang, Xiaoke Peng, Xuehui Yu, Qiang Wang, Junliang Xing, Guorong Li, Guodong Guo, Qixiang Ye, Jianbin Jiao, et~al.
\newblock Anti-uav: a large-scale benchmark for vision-based uav tracking.
\newblock \emph{IEEE Transactions on Multimedia}, 25:\penalty0 486--500, 2021.

\bibitem[Li et~al.(2023)Li, Yuan, Sun, Wang, Liu, and Liu]{li2023global}
Yifan Li, Dian Yuan, Meng Sun, Hongyu Wang, Xiaotao Liu, and Jing Liu.
\newblock A global-local tracking framework driven by both motion and appearance for infrared anti-uav.
\newblock In \emph{Proceedings of the IEEE/CVF Conference on Computer Vision and Pattern Recognition}, pages 3025--3034, 2023.

\bibitem[Xiao and Feroskhan(2022)]{9764628}
Jiaping Xiao and Mir Feroskhan.
\newblock Cyber attack detection and isolation for a quadrotor uav with modified sliding innovation sequences.
\newblock \emph{IEEE Transactions on Vehicular Technology}, 71\penalty0 (7):\penalty0 7202--7214, 2022.

\bibitem[Xiao and Feroskhan(2024)]{10439240}
Jiaping Xiao and Mir Feroskhan.
\newblock Learning multi-pursuit evasion for safe targeted navigation of drones.
\newblock \emph{IEEE Transactions on Artificial Intelligence}, pages 1--14, 2024.

\bibitem[Xiao et~al.(2023)Xiao, Zhang, Zhang, and Feroskhan]{xiao2023vision}
Jiaping Xiao, Rangya Zhang, Yuhang Zhang, and Mir Feroskhan.
\newblock Vision-based learning for drones: A survey.
\newblock \emph{arXiv preprint arXiv:2312.05019}, 2023.

\bibitem[Xiao et~al.(2024)Xiao, Chee, and Feroskhan]{10417793}
Jiaping Xiao, Jian~Hui Chee, and Mir Feroskhan.
\newblock Real-time multi-drone detection and tracking for pursuit-evasion with parameter search.
\newblock \emph{IEEE Transactions on Intelligent Vehicles}, pages 1--11, 2024.

\bibitem[Yuan et~al.(2024)Yuan, Yang, Nguyen, Nguyen, Yang, Liu, Li, Wang, and Xie]{yuan2024mmaud}
Shenghai Yuan, Yizhuo Yang, Thien~Hoang Nguyen, Thien-Minh Nguyen, Jianfei Yang, Fen Liu, Jianping Li, Han Wang, and Lihua Xie.
\newblock Mmaud: A comprehensive multi-modal anti-uav dataset for modern miniature drone threats.
\newblock \emph{arXiv preprint arXiv:2402.03706}, 2024.

\bibitem[Zhao et~al.(2022)Zhao, Zhang, Li, and Wang]{zhao2022vision}
Jie Zhao, Jingshu Zhang, Dongdong Li, and Dong Wang.
\newblock Vision-based anti-uav detection and tracking.
\newblock \emph{IEEE Transactions on Intelligent Transportation Systems}, 23\penalty0 (12):\penalty0 25323--25334, 2022.

\end{thebibliography}
}


\end{document}